\documentclass[letterpaper]{article} 
\usepackage{aaai25} 
\usepackage{times}  
\usepackage{helvet}  
\usepackage{courier}  
\usepackage[hyphens]{url}  
\usepackage[pdftex]{graphicx} 
\def\UrlFont{\rm}  
\usepackage{natbib}  
\usepackage{caption} 
\usepackage{algorithm}
\usepackage{algorithmic}
\usepackage{todonotes}
\usepackage{cleveref} 
  \crefname{section}{Sect.}{Sects.}
  \Crefname{section}{Section}{Sections}
  \crefname{figure}{Fig.}{Figs.}
  \Crefname{figure}{Figure}{Figures} 
  \crefname{definition}{Definition}{Definitions}
  \crefname{equation}{}{}
  \Crefname{equation}{Equation}{Equations}
  \crefname{table}{Tab.}{Tabs.}
  \Crefname{table}{Table}{Tables}

\usepackage{tabularx}
\usepackage{booktabs}
\usepackage{lipsum}

\usepackage{newfloat}
\usepackage{listings}
\DeclareCaptionStyle{ruled}{labelfont=normalfont,labelsep=colon,strut=off} 
\floatstyle{ruled}
\newfloat{listing}{tb}{lst}{}
\floatname{listing}{Listing}
\title{Towards Automated Safety Requirements Derivation Using Agent-based RAG}

\author {
    Balahari Vignesh Balu,
    Florian Geissler,
    Francesco Carella,
    Joao-Vitor Zacchi,\\
    Josef Jiru,
    Nuria Mata,
    Reinhard Stolle
}
\affiliations{
    Fraunhofer IKS, Fraunhofer Institute for Cognitive Systems IKS, Munich, Germany\\
    \{balahari.balu, florian.geissler, francesco.carella, 
    joao-vitor.zacchi, josef.jiru, nuria.mata, reinhard.stolle\}@iks.fraunhofer.de
}

\begin{document}

\maketitle
%
\begin{abstract}
We study the automated derivation of safety requirements in a self-driving vehicle use case, leveraging LLMs in combination with agent-based retrieval-augmented generation.
Conventional approaches that utilise pre-trained LLMs to assist in safety analyses typically lack domain-specific knowledge. Existing RAG approaches address this issue, yet their performance deteriorates when handling complex queries and it becomes increasingly harder to retrieve the most relevant information. This is particularly relevant for safety-relevant applications.
In this paper, we propose the use of agent-based RAG to derive safety requirements and show that the retrieved information is more relevant to the queries. We implement an agent-based approach on a document pool of automotive standards and the Apollo case study, as a representative example of an automated driving perception system.
Our solution is tested on a data set of safety requirement questions and answers, extracted from the Apollo data.
Evaluating a set of selected RAG metrics, we present and discuss advantages of a agent-based approach compared to default RAG methods.
\end{abstract}
\section{Introduction}

Safety analysis processes are time-consuming and require involvement of experts who are well versed in the relevant domain-specific knowledge. Large language models (LLMs) might assist in some repetitive tasks, like deriving or aligning requirements or populating templates for safety documentation, while keeping the output language (keywords, quality) consistent. LLMs can also perform reviews of existing safety documentation, checking consistency and completeness, assuring traceability between requirements, system elements and verification steps. However, to efficiently support safety engineers in these tasks, LLMs must generate highly reliable and explainable results.

At its core, the ability of LLMs to process natural language is enabled by an attention mechanism that identifies semantic correlations between tokens to find meaningful continuations \cite{Vaswani2017}. Nevertheless, there is always the possibility that plausible continuations are not factually grounded, leading to so-called "hallucinations". These events have to be suppressed or mitigated efficiently in order to make LLMs a truly useful tool for safety analysis tasks.

The risk of hallucinations is related to the information accessible to the LLM, and its suitability to solve the task at hand.
Model weights are typically trained on a large corpus of data representing general knowledge.
To integrate additional domain-specific information, techniques such as fine-tuning or retrieval-augmented generation (RAG) \cite{lewis2020retrieval} are commonly used. RAG approaches may reduce the chance of hallucinations if they succeed in retrieving relevant pieces of information, which then serve as context in the final LLM prompt. On the other hand, if irrelevant or even incorrect context is retrieved, the probability of hallucinations is exacerbated. 

It is therefore of great interest for all safety analysis tasks to identify efficient and reliable RAG architectures.
Among various proposed methods \cite{Ranjan2024}, multi-agent RAG systems \cite{Talebirad2023, Salve2024} seek to create a fruitful interplay of agents which each have the capability to access an individual set of documents via RAG.
In this paper, we study how a customized agent-based RAG approach fares on a problem encountered in almost any safety engineering workflow: deriving safety requirements for a given system component with a known insufficiency. 
Our work provides important observations on how RAG-based LLM systems can be leveraged for safety analysis processes.

\begin{figure*}[h]
    \centering
    \includegraphics[width = 0.9\textwidth]{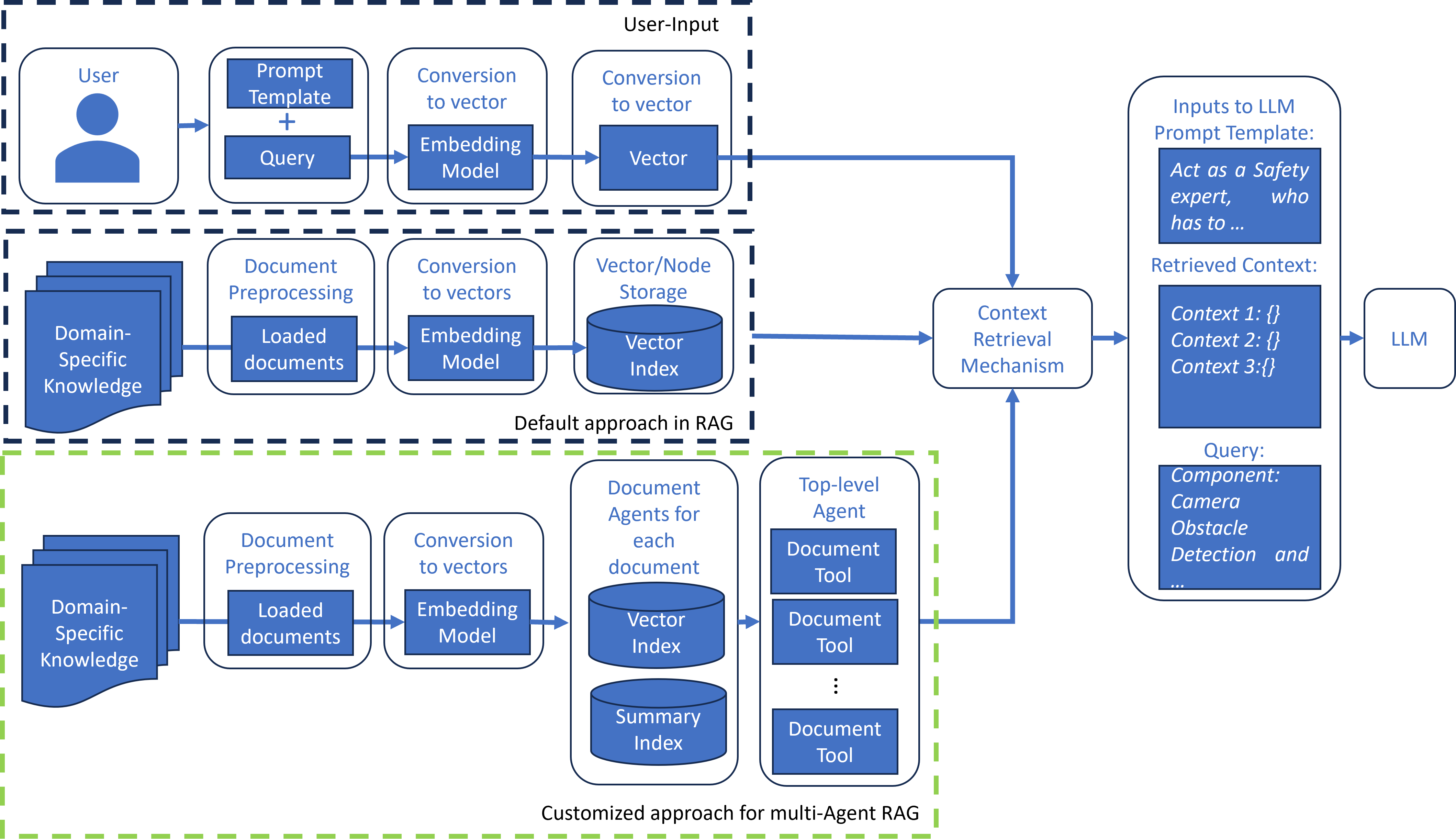}
    \caption{A RAG pipeline showcasing how domain knowledge is pre-processed and stored to be retrieved later as context, serving as input to the LLM together with the user query. The  default (conventional) RAG approach is replaced by a customized approach highlighted in green, which enables a refined context retrieval mechanism illustrated in detail in Fig.~\ref{fig:agent_based_rag}.}
    \label{fig:conventional_vs_agent_based_rag}
\end{figure*}
\section{Background}
\begin{figure*}[h]
    \centering
    \includegraphics[width = 0.8\textwidth]{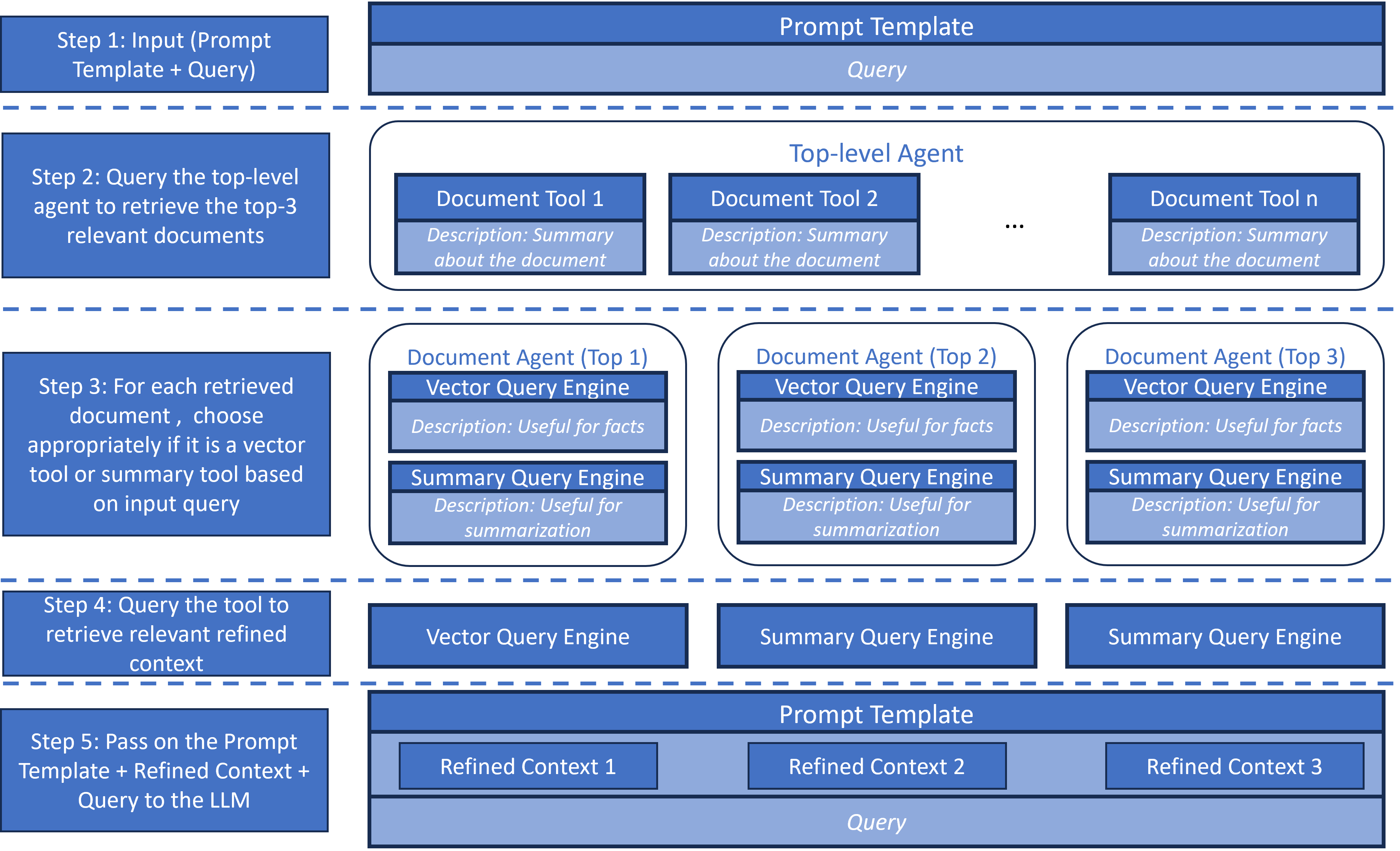}
    \caption{The sequence of steps (read from top-to-bottom) depicting the workflow of the proposed alternate method of context retrieval mechanism to retrieve refined contexts in the  agent-based RAG approach proposed in this paper.}
    \label{fig:agent_based_rag}
\end{figure*}

\textbf{General workflow of deriving safety requirements:}
The task of deriving safety requirements in the automotive domain is typically based on key standards such as ISO 26262 \cite{iso26262}, which has a focus on functional safety to ensure safety by mitigating system failures (hardware/software faults), and ISO 21448 (SOTIF) \cite{ISO21448}, that discusses the safety of the intended functionality by addressing performance limitations and misuse scenarios when no faults exist.
The ISO 26262 provides guidelines for the complete safety lifecycle of automotive systems. The foundations for deriving the safety requirements  is grounded in the safety lifecycle during the concept phase in a sequential manner - 1) System definition, 2) hazard analysis, 3) safety goals and requirements. The process begins with the item definition that contains the different functions, interfaces, and boundaries of the system and its operating environment with other systems and users. This task primarily describes the system's purpose and context. 

After defining the system, its modes of operation and operational design domain (ODD) are used to identify the operational scenarios. The subsequent hazard analysis is based on the Hazard and Risk Analysis (HARA) as outlined by automotive standards \cite{iso26262, ISO21448}. It is a structured process specific to each domain and is used to assess the hazards, associated risks and required safety measures and serves as a prerequisite for the elicitation of safety requirements. The goal is to identify hazardous situations that can cause harm. 
Techniques like hazard and operability analysis (HAZOP) are used to identify possible deviations from an intended operation to deduce hazardous scenarios.
Each of these hazardous events are assessed for associated risks based on three parameters - severity, exposure and controllability. Based on the values assigned, automotive safety integrity level (ASIL) ratings are determined.
The HARA report documents the hazards, levels of risks associated with it and the ASIL classification.
For every hazardous event identified, a safety goal is specified so that the associated risk is addressed or controlled. Among those, similar safety goals are aggregated and the combined goal takes the highest ASIL rating. 
High-level safety requirements are defined to meet these safety goals. Since these are referring to the entire system, they need to be translated to specific subsystems or individual components with the help of techniques like a fault tree analysis (FTA). The requirements for specific components can help to prevent situations like the failure of a component, or the corruption or loss of messages during the communication of components.

The above tasks are performed by teams of various stakeholders and safety experts. The nature of these tasks typically requires human brainstorming, expertise with previous domain knowledge, and compliance with relevant standard guidelines. It is also necessary to understand the outputs generated at each stage, as they represent essential input to the next stage. In short, the various tasks in safety analysis workflow relies on the intricate, domain-specific experience held by safety experts, including a familiarity with the relevant standards. A LLM assistant should be able to mimic such characteristics that are expected from a safety expert.

\textbf{Retrieval Augmented Generation:}
LLMs achieve state-of-the-art results for domain-specific NLP tasks typically only upon fine-tuning or respective re-training. However, if the fine-tuning training set includes low-quality or too few samples, the answer quality can still be poor. Pre-trained LLMs can further generate responses with factual inaccuracies (or hallucinations). RAG models, as introduced in \cite{lewis2020retrieval}, often provide a cost-effective solution to the aforementioned problem. Hereby, external information gets embedded and stored in a vector store. When paired with a pre-trained retriever with access to the stored knowledge, the RAG model is able to generate outputs grounded in the provided knowledge, while still holding the ability to be diverse, factual and specific in nature. 
Various approaches \cite{izacard2020leveraging,guu2020retrieval, borgeaud2022improving, izacard2023atlas} have emerged over time that seek to retrieve more targeted and relevant context, as well as to improve the retrieval quality \cite{asai2023selfraglearningretrievegenerate}.


\section{Related Work}
\label{sec:relatedwork}

The use of LLMs to assist in Systems Theoretic Process Analysis (STPA) was studied by the authors of \cite{qi2023safetyanalysiseralarge}. They used LLMs to perform STPA and derive a set of safety requirements for two case studies - automated emergency braking and electricity demand side management systems. Different collaboration schemes between human experts and LLMs were explored. The authors also analyze the effects of input complexities and prompt engineering on the LLM results. The paper emphasizes the need for human safety experts to complement the safety analysis by LLMs, while highlighting the issue of non-reliability of LLMs, i.e., the generation of factually inaccurate and inconsistent outputs due to their limited domain-specific knowledge.

The authors of \cite{nouri2024engineeringsafetyrequirementsautonomous} explore the use of LLMs to automate the HARA, where they define a LLM-based pipeline and try to improve it by incorporating feedback from safety experts. 
The limitation of a lack of domain-specific knowledge in understanding technical terms and processes within automotive and safety domains is acknowledged, along with other difficulties such as its limited capability in interpreting technical information presented as figures. 
Further, inconsistencies are identified in the generated set of safety requirements in relation to the use of modal verbs such as \textit{"should"} and \textit{"shall"}, which has specific meaning in the context of automotive safety. 
The authors countered these identified limitations by breaking down the HARA into smaller tasks developing specific context, guides and prompts for each identified sub-tasks. To address hallucinations and improved explainability, such background information is included within the prompts to stimulate reasoning along with generated outputs.

In \cite{geissler2024conceptguidedllmagentshumanai}, a hybrid LLM-classical safety workflow is discussed for scenarios such as fault propagation in a system graph. A LLM agent is first guided along pre-formulated categories to identify the specific problem at hand represented by the input prompt. Subsequently, deterministic tools are leveraged to actually solve the task, using variables prepared by the previous LLM analysis. 

Another relevant study is presented in 
\cite{sivakumar2024prompting}. Here, the authors assess the ability of LLMs to understand and generate safety cases based on the Goal Structuring Notation (GSN). They tested the correctness of generated safety cases by comparing them against ground truth safety cases derived from other reference research papers. Apart from concerns such as hallucinations, or lexical correctness of generated content, the findings underscore that the unavailability of contextual information about reference safety cases limits the scope and comprehensiveness of the generated safety cases. While the previous work focused on the generation of the safety case, the authors of \cite{gohar2024codefeaterusingllmsdefeaters} use LLMs to identify potential defeaters in claims that are part of an assurance case. They iteratively process each claim with role-based prompting to demonstrate the LLM's effectiveness in identifying potential defeaters in a zero-shot setting.

While these works primarily explore the intersection between LLMs and safety analysis processes, most of them focus on zero-shot prompting approaches or improvised prompt engineering techniques to investigate the use of an LLM for a particular tasks. Frequently encountered concerns include hallucinations and the limited availability of domain-specific knowledge for the LLMs. Our work takes a step further in addressing this concern by proposing a agent-based RAG-based approach to effectively tailor the support to the safety analysis process.

\section{Methodology and Experimental Setup} \label{sec:experiments}

\begin{table}

\begin{tabularx}{\columnwidth}{@{}XX@{}}
    \toprule
    \textbf{Metric} & \textbf{Description}\\\midrule
     \textbf{Normalized answer similarity score (NASS)} & Similarity of reference answer and LLM answer, as assessed by a LLM judge. \\ \midrule
    \textbf{Retrieval precision (RP)} & Decide whether the retrieved context is relevant to the question, as assessed by a LLM judge. \\ \midrule
    \textbf{Augmentation accuracy (AA)} & Decide whether the retrieved context is used in the LLM answer, as assessed by a LLM judge. \\ \midrule
    \textbf{Augmentation precision (AP)} & Decide whether the relevant retrieved context used in the LLM answer, using RP, AA.  \\ \midrule
    \textbf{Answer consistency (AC)} & Summarizes main points from the LLM answer and checks which portion of those main points is found also in the context, as assessed by a LLM judge. \\\bottomrule
\end{tabularx}
\caption{Selected metrics adapted from \cite{tonicai_git}. All metrics are normalized to a range of $0-1$.}
\label{tab:metrics_summary}
\end{table}

\textbf{Customized approach for agent-based RAG:}
To facilitate the incorporation of domain-specific knowledge while avoiding the overhead cost of fine-tuning or retraining of LLMs for specific tasks, we propose a novel approach that empowers the default RAG-based approach, utilizing document agents to perform tasks with higher complexities. This strategy enhances reliability by working within the boundaries of provided domain knowledge and, as a consequence, mitigates hallucinations. Furthermore, the use of document agents allows for a modular and scalable structure to accommodate  domain-specific knowledge that might be constantly updated. 


The workflow of a default RAG-based approach is showcased in Fig.~\ref{fig:conventional_vs_agent_based_rag}. 
The domain-specific knowledge in the form of documents is transformed into the vector representations using the embedding models, preserving their semantic meaning.
The relevant documents of different formats are parsed into different chunks. These chunks are converted into different embeddings along with the metadata, for use by the LLM. The embeddings are stored in the form of vector representations in the \textit{Vector Store Index} to be retrieved later. The latter is performed by a \textit{Query Engine}. It acts as the interface between the AI model and the indices enabling the retrieval task and translates the query into a suitable format for the indices to be retrieved. During retrieval, the query embedding is compared with the stored chunks using a similarity metric to identify those chunks that are relevant to the query. Finally, the retrieved chunks are included as context along with the query and processed by the LLM to generate a response.

In addition to the components of a default RAG-based pipeline, 
the \textit{Document Agent} is introduced for different sets of documents. As shown in the step 3 of Fig.~\ref{fig:agent_based_rag}, each document agent represents a query engine that contains a \textit{Vector Store Index} and a \textit{Summary Index}. Along with the vector store index, the summary index introduced in \cite{summaryindex_llamaindex} is an iterative, hierarchical method for summarizing the document, where a tree-like structure is created with summaries stored at lower and higher levels, respectively. This enables the handling of long, complex documents while allowing for the storage of abstracted information contained in those documents.
A query engine is defined for each of these indices. At the time of creation, an appropriate description is passed, such that the vector query engine would be used to answer facts about the document and the summary query engine would be used to answer summarization questions about the document. Whenever the document gets chosen to retrieve relevant context, based on the query, either vector query engine or summary query engine will be invoked. This selection of an appropriate query engine is implemented with the OpenAI agent framework \cite{openai_agent_framework}. The result is a document agent for that particular document.
While the document agent helps to choose between vector or summary query engine of a document, choosing the appropriate document is facilitated by the \textit{Document Tool}.
The document agents for each document are wrapped with another query engine along with a short description (1-2 lines) about the contents of the document. Multiple document tools are indexed together to form the \textit{Top-level Document Agent}. 

 The top-level document agent gets invoked upon receiving a query from the user, as shown in step 2 of Fig.~\ref{fig:agent_based_rag}. This retrieves the top-3 relevant documents. The same query is again used within the document agent representing the top-3 retrieved documents to choose between a vector query engine or a summary query engine. After choosing the appropriate query engine, a refined context is produced from each of the top-3 documents, if the documents do not contain any relevant information, then the context gets discarded. These refined contexts are passed on along with the user input to the LLM to generate the final output.




\textbf{Dataset:} To evaluate the methodology, we chose the publicly available Apollo use case study for an automated driving perception system \cite{apollo_data, apollo_git, Kochanthara2024}. 
A dataset of $58$ question-answer pairs ($30$ safety tactic-based and $28$ ML-design components) was collected from the "Design Assessment" section of \cite{apollo_data}, representing a list of critical system components and their respective safety requirements, as generated by safety experts.
In our experimental setup, a ground truth reference answer is given by a safety requirement, and the respective question is a concatenation of a component or a set of components (also called "pipeline"), as well as a known insufficiency, and possibly a trigger condition, if provided in ~\cite{apollo_data}.
Every question is further embedded in a system prompt. Following best prompt engineering practices \cite{Zhou2023}, the system prompt was crafted to be as specific as possible, eliminating ambiguity regarding the task. A single data point of our dataset is then, for example (see also Tab.~\ref{tab:example_raw_data}):
\begin{quote}
\it
\textbf{Prompt:} "Act as a safety engineer, who has the task to derive safety requirements          for a given component pipeline.
As input, you are given the pipeline, a known potential functional insufficiency, and possibly a trigger condition.
\newline
Output a safety requirement, i.e. a description how the function of the component pipeline shall not perform in case the known insufficiency occurs. 
Consider the function of the component pipeline and possible further downstream system functions to state what shall not happen in case of the functional insufficiency.
Keep your answer as brief as a single sentence, but make sure a system-specific requirement is given.
Begin your statement with 'If...'
\newline
INPUT: {///}\{Question\}{///}
\newline
OUTPUT:", 
\newline
\textbf{Question:} "Pipeline: Camera obstacle detection, classification, and tracking pipeline, Known potential function insufficiency: deteriorated performance of camera based object detection and tracking due to adverse weather conditions , Trigger condition: moderate increment levels of rain"
\newline
\textbf{Reference answer:} "If the performance of Camera obstacle detection, classification, and tracking pipeline is deteriorated due to moderate increment levels of rain, then this deterioration in performance shall not lead to an incorrect estimation of the state of vehicles or other obstacles."
\end{quote}

The task given to the LLM is to generate safety requirements for the given inputs, and the performance with respect to the ground truth is evaluated using the metrics discussed in the next section.
The entire Apollo documentation \cite{apollo_data} was used as source material for RAG (Retrieval-Augmented Generation) to assist in this task, except for the safety requirement documentation containing the ground truth. Furthermore, we added content adopted from relevant automotive standards such as ISO PAS 8800 \cite{ISO8800}, ISO 26262 \cite{iso26262}, ISO 21448 (SOTIF) \cite{ISO21448}, ISO/TR 4804 \cite{ISOTR4804} and UN157 \cite{no2022157} to the document pool accessible by RAG.

\textbf{Model:} For LLM access, a pre-trained GPT-3.5 model \cite{OpenAI} was used both to generate predictions and to act as a judge LLM for the respective metrics. For both purposes, the default temperature was used.
We compare a standard RAG approach, our proposed agent-based RAG, and single LLM calls without RAG context. To implement both the default and our advanced RAG pipeline, we made use of the LLamaIndex \cite{llamaindex} library as well as the FAISS vector store \cite{douze2024faiss}. The default chunk size was chosen for parsing context in the vector store.

\textbf{Metrics:} In order to assess the quality of the different approaches, we apply the metrics in Tab.~\ref{tab:metrics_summary}, see \cite{tonicai_git}.
Each metric are applied to a subset of an individual tuple of {\it (question, retrieved context, LLM answer, reference answer)}.
In particular, note that RP directly evaluates the quality of the context retrieval step, while AA, AP, AC are focused on a semantic match between context and the answer.

\section{Results}
\label{sec:results}

\begin{figure}[H]
    \centering
    \includegraphics[width=0.48\textwidth]{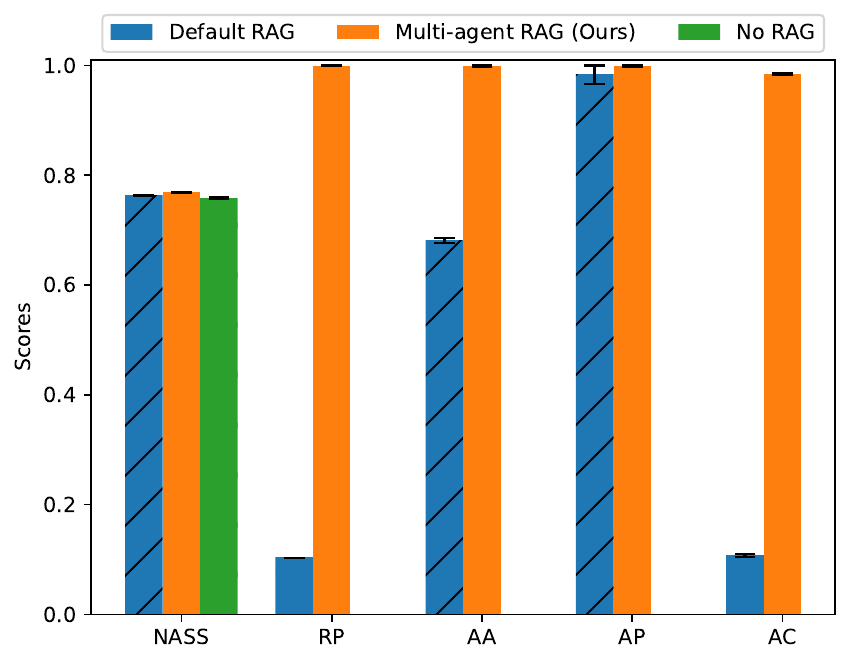} 
    \caption{Average and standard error for the metrics selected in Tab.~\ref{tab:metrics_summary} for $10$ independent runs.}
    \label{fig:metrics_plot} 
\end{figure}

\begin{table*}
    \centering
    \small
    \begin{tabular}{p{3cm}p{12cm}}
        \toprule
        Prompt & 
        "Act as a safety engineer, who has the task to derive safety requirements          for a given component pipeline.
        As input, you are given the pipeline, a known potential functional insufficiency, and possibly a trigger condition.
        Output a safety requirement, i.e. a description how the function of the component pipeline shall not perform in case the known insufficiency occurs. 
        Consider the function of the component pipeline and possible further downstream system functions to state what shall not happen in case of the functional insufficiency.
        Keep your answer as brief as a single sentence, but make sure a system-specific requirement is given.
        Begin your statement with 'If...'
        \newline
        INPUT: {///}Pipeline: Camera obstacle detection, classification, and tracking pipeline, Known potential function insufficiency: deteriorated performance of camera based object detection and tracking due to adverse weather conditions , Trigger condition: moderate inclement levels of rain{///}
        \newline
        OUTPUT:" \\
        \hline
        Reference answer & "If the performance of Camera obstacle detection, classification, and tracking pipeline is deteriorated due to moderate inclement levels of rain, then this deterioration in performance shall not lead to an incorrect estimation of the state of vehicles or other obstacles." \\
        \hline
        LLM answer \newline (no RAG) & "If the camera based object detection and tracking performance deteriorates due to moderate inclement levels of rain, the pipeline shall not provide inaccurate or delayed obstacle detection, classification, and tracking." (NASS: 4.5/5) \\ 
        \hline
        LLM answer (Default RAG) & "If the performance of the camera-based object detection and tracking deteriorates due to moderate inclement levels of rain, the camera obstacle detection, classification, and tracking pipeline shall not rely solely on the camera input and shall activate an alternative sensor for object detection and tracking." (NASS: 3/5)\\
        \hline
        LLM answer (Agent-based RAG) &  "If the camera-based object detection and tracking performance deteriorates due to moderate inclement levels of rain, the pipeline shall not rely solely on camera data for obstacle detection and tracking, but shall also utilize other sensor inputs such as radar or lidar." (NASS: 4.5/5)\\
        \bottomrule
    \end{tabular}
    
    \caption{Selected example LLM answers to the same prompt for each studied RAG approach.}
    \label{tab:example_raw_data}
\end{table*}

\begin{table*}
    \centering
    \small
    \begin{tabular}{p{2cm}p{14cm}} 
        Approach  & Metric components \\
        \toprule
       Default RAG & \vspace{-\baselineskip} \begin{itemize}
            \item Retrieved context:
            \begin{itemize}
                \item
                "ISO 21448:2022(E) Figure 14 2014 Example of system architecture with the fusion of two diverse sensors. The classification can also be used during the definition of the validation strategy, where the validation targets for multiple-point functional insufficiencies can be reduced subject to independence considerations (see Clause 9  and C.6.3 ). ..." ,
                \item
                "ISO 21448:2022(E) 2014 known potential functional insufficiencies of the system and its elements and known potential triggering conditions (including reasonably foreseeable direct misuse) that could lead to a hazardous behaviour based on external information or lessons learnt (e.g. 13.5 ). ...",
                \item
                "ISO 21448:2022(E) Key a Depending on the architecture of the system this functional insufficiency on an element level can be recognized either as a single-point functional insufficiency  (3.28 ) or a multiple point functional insufficiency  (3.19 ). b An output insufficiency, either by itself or in combination with one or more output insufficiencies of other elements, ..."
            \end{itemize}
            \item Answer consistent with context: True
            \item Context relevancy: False
            \item Main points: 
            \begin{itemize}
                \item "Performance of camera-based object detection and tracking deteriorates in rain",
                \item
                "Camera obstacle detection, classification, and tracking pipeline should not rely solely on camera input",
                \item
                "Activate alternative sensor for object detection and tracking"
            \end{itemize}
            \item Statement derived from context: True
            \item Main points from context: False, False, False
        \end{itemize} \\ \hline
      Agent-based RAG & 
      \vspace{-\baselineskip} \begin{itemize}
            \item (Summarized) Retrieved context:
            \begin{itemize}
                \item "If the camera-based object detection and tracking performance deteriorates due to moderate inclement levels of rain, the pipeline shall not rely solely on camera data for obstacle detection and tracking, but shall also utilize other sensor inputs such as radar or lidar."
                \item
                "If the camera-based object detection and tracking performance deteriorates due to moderate inclement levels of rain, the pipeline shall not rely solely on camera data for obstacle detection and tracking, but instead utilize additional sensor inputs or alternative detection methods."
            \end{itemize}
            \item Answer consistent with context: True
            \item Context relevancy: True
            \item Main points: 
            \begin{itemize}
                \item "Camera-based object detection and tracking performance deteriorates in moderate rain"
                \item
                "Pipeline should not rely solely on camera data for obstacle detection and tracking"
                \item
                "Other sensor inputs such as radar or lidar should be utilized"
            \end{itemize}
            \item Statement derived from context: True
            \item Main points from context: True, True, True
        \end{itemize} 
        \vspace{-\baselineskip} \\
        \bottomrule
    \end{tabular}
    
    \caption{Traced context and analyzed components of the metrics RP, AA, AP, AC using the example given in Tab.~\ref{tab:example_raw_data}. For the default RAG, the retrieved context is shortened for visibility. For the agent-based RAG, the context is summarized according to the agent's instructions.}
    \label{tab:example_raw_data_context}
\end{table*}

The Fig.~\ref{fig:metrics_plot} shows the result of evaluating the safety requirements dataset with our agent-based RAG approach, comparing to the two baselines of a single LLM call (no RAG, only for NASS), and a default RAG pipeline.
We make the following observations.
\begin{itemize}
\item The NASS score is very similar for both RAG approaches and even LLM calls without RAG. This suggests that, on average, the LLM can leverage generic knowledge to phrase answers that are assessed as similar to the correct response. We further verified that alternative similarity metrics, such as the BERT score \cite{zhang2020bertscoreevaluatingtextgeneration}, exhibit only minimal variations for the three approaches.
An inspection of selected examples (see Tab.~\ref{tab:example_raw_data}) indicates, however, that RAG-assisted answers do contain additional information that a human user would deem useful. For example, for deteriorated performance of the camera system in case of rain, the LLM without RAG states that this should not lead to an incorrect estimation of the state of other vehicles or obstacles. The answer with default RAG further mentions that alternative sensors should be activated. With the agent-based RAG approach, the LLM answer additionally refers to such other sensor modalities like radar or lidar that are available within the architecture of Apollo, demonstrating the ability to identify nuances in the process of safety requirements derivation.
Nevertheless, all answers receive a similar NASS score by the LLM judge.
We therefore conclude, that the similarity metrics such as NASS or BERT score are not able to capture subtle details that might still be of interest to a human reviewer.
We expect that the NASS score will be more discriminative in more complex problems, or when using smaller LLM models with less generic knowledge.
\item 
The RP metric is significantly higher for our agent-based RAG approach than for the default RAG. In the former, the context used for evaluation is information that is pre-processed by the RAG agents after individual calls. Such a multi-step approach is able to condense chunks from original source documents into compact summaries, that are semantically highly relevant for the query. This is not the case for default RAG, where only raw context chunks are retrieved from the source documents. This can have the effect, that the chunks retrieved by default RAG match only individual keywords from the prompt (for example "insufficiencies", standard names, etc.) while missing the overall meaning. We further observe, that the default RAG is prone to getting "stuck" with a specific document, i.e., the retriever extracts all top chunks from the same, often not even most significant document. As the low RP score indicates, it tends to retrieve plenty of irrelevant information chunks, see also Tab.~\ref{tab:example_raw_data_context}. On the other hand, the agent-based RAG is more successful in fusing information from different document sources.
\item 
Due to the above-mentioned condensation of the retrieved contexts, the final LLM answers are very similar to the retrieved contexts in the agent-based approach. This is reflected in a high AA, AP score, as well as the answer consistency. Default RAG also scores high in AP, suggesting that if relevant context was retrieved, this also carries over to the LLM answer. However, the answers typically contain content that is missing in the context (see low AC, AA), and thus had to be generated from generic knowledge. The agent-based RAG already eliminates irrelevant context during pre-processing, leading to very high AA, AP, AC scores.
\end{itemize}

\section{Conclusion and Outlook}
\label{sec:conclusion}
For the given safety requirements data set, we find that LLM answers with high similarity to the reference answer can be generated without RAG, or using default RAG. Inspecting further metrics such as the RP, AA and, AC, however, we see that our agentic RAG approach retrieves significantly more relevant context than the default RAG. 
We interpret these results such that the problem setup (used data set and model) is likely too simple to reveal the full benefit of the agentic RAG approach: Good solutions can be found simply by leveraging generic knowledge of the LLM, such that the quality of the retrieved context has no significant impact on the answer similarity alone. Nevertheless, the fact that this strategy retrieves a measurably more relevant context is very promising and insightful for safety-critical applications. A closer inspection also shows, that a LLM-based similarity score (here NASS) does not necessarily capture all quality attributes of a generated requirement. Although a verification with human expert reviews was out of scope of this paper, we suspect that such a human scoring metric would prefer the agentic RAG over the default RAG due to its higher density of potentially useful details. 
As next steps, we will test our strategy with more complex data sets or simpler LLM models, in order to establish a closer connection to the retrieved context relevance and the LLM answer quality.

\section{Acknowledgments}
This work was funded by the Bavarian Ministry for Economic
Affairs, Regional Development and Energy as part of
a project to support the thematic development of the Institute
for Cognitive Systems.

\bibliography{bibliography}

\end{document}